\documentclass[11pt]{article}
\usepackage[T1]{fontenc}
\usepackage[letterpaper,margin=1in]{geometry}
\usepackage{lmodern}
\usepackage{microtype}
\usepackage{amsmath,amssymb}
\usepackage{graphicx}
\usepackage{longtable,booktabs,array,calc}
\usepackage[colorlinks=true,allcolors=blue]{hyperref}
\providecommand{\tightlist}{\setlength{\itemsep}{0pt}\setlength{\parskip}{0pt}}
\title{Verification Behavior Across Five Qwen2.5-Instruct Checkpoints}
\author{YaJie Yin}
\date{}
\hypersetup{pdftitle={Verification Behavior Across Five Qwen2.5-Instruct Checkpoints},pdfauthor={YaJie Yin}}
\begin{document}
\maketitle
\begin{abstract}
We evaluate five Qwen2.5-Instruct checkpoints (0.5B to 32B) under four prompting conditions using three fixed tasks per information condition and five generations per task. Two AI raters scored 600 trajectories. Under the coordinated instruction, both raters recorded strict declarations of insufficiency in 11/15 cases at 7B, 0/15 at 14B, and 5/15 at 32B. The runner sent a standard follow-up when a request detector fired. The three checkpoints received at least one follow-up in 15, 12, and 15 cases, respectively. On information-sufficient tasks at 3B through 32B, scored format markers appeared in 60/60 protocol trajectories and 0/60 baseline trajectories. After adjudication, 1/60 in each condition met the diagnostic-execution criterion (H3). Across all checkpoints and prompts, 7/300 information-sufficient trajectories met H3 after adjudication, all on the same task. The prespecified upward transition did not occur in this run. The results separate visible structure, diagnostic execution, and persistence after a request for evidence.
\end{abstract}
\section{Introduction}\label{introduction}

Language models are often asked to verify analyses, audit claims, or judge whether available evidence supports a conclusion. We count three observable behaviors: the presence of a format marker, the execution of a diagnostic check under the rubric, and continued requests for missing evidence after a follow-up. A response can also make a useful judgment below the diagnostic-execution criterion, as one case in the discussion illustrates.

This paper asks whether a verification protocol produces increasingly substantive behavior across sizes of a single instruction-tuned model family. We compare five Qwen2.5-Instruct checkpoints with four instructions and six fixed tasks. Three tasks provide enough information to evaluate a flawed report. The other three withhold information needed to answer. This design distinguishes the execution of a check from the decision to withhold judgment, and measures both alongside visible protocol structure.

Under the coordinated instruction, strict declarations vary non-monotonically across checkpoints. At 3B through 32B, all protocol trajectories show a scored marker and few meet the diagnostic-execution criterion. The five predictions cover the anticipated transition and the direction of a historical 14B comparison. A companion study uses a related evidence-sufficiency task family and measures final boundary-keeping under action pressure following a training intervention~\cite{yinboundary2026}.

\section{Related work}\label{related-work}

Studies of chain-of-thought prompting have found size-dependent task-accuracy gains \cite{wei2022}. The Qwen2.5 family provides instruction-tuned checkpoints at multiple sizes \cite{qwen2025}. We examine whether a structured prompt changes the verification behavior of these checkpoints on the same tasks and separate protocol headings from the execution of a check.

Zhao et al. assess how individual chain-of-thought steps affect the final answer in their experiments \cite{zhao2025}. Our unit of observation is a complete model response or interaction, and our rubric uses an observable action or conclusion as evidence. Seo et al. evaluate fact verifiers and show how ambiguity and annotation quality can affect comparative results \cite{seo2025}. The distinction between an expressed verification verdict and the evidence supporting it also motivates Verification Autonomy Levels \cite{yin2026}. Here that distinction motivates separate measures of reporting format, substantive checking, and handling of insufficient evidence.

\section{Methods}\label{methods}

\subsection{Within-family scale scan design}\label{within-family-scale-scan-design}

We evaluated five Qwen2.5-Instruct checkpoints (0.5B, 3B, 7B, 14B, and 32B) using the same six tasks and four prompting conditions. The information-sufficient tasks (T8-\/-T10) asked the model to check a supplied claim, and the information-insufficient tasks (T18-\/-T20) withheld evidence needed for a judgment. The conditions were a baseline task prompt, the seven-step verification protocol, a coordinated protocol with an explicit sequence for requesting missing information, and an instruction explicitly requiring a declaration of insufficiency. Each task-\/-condition combination was sampled five times per checkpoint. The 600 cases comprise 300 cases per information condition and 15 cases per checkpoint, prompt, and information condition.

All checkpoints ran under vLLM 0.25.1 on a single 32 GB GPU, with temperature 0.4, top-p 0.95, and a 2,048-token maximum per response. The 0.5B through 14B checkpoints used BF16, and the 32B checkpoint used AWQ quantization. The 14B deployment used a maximum context length of 5,120 tokens and a GPU memory utilization setting of 0.97. The other deployments used a context limit of 8,192 tokens, and the 32B deployment used a utilization setting of 0.95. Tasks, conditions, and repeats were processed in fixed blocks without a separate warm-up. Repeat identifiers 1-\/-5 were also passed as seeds. Replay checks showed that identical seed values did not ensure token-level reproducibility in this runtime, so the analysis uses aggregate response counts. Checkpoint size co-varies with quantization or deployment settings at some points.

The runner allowed up to three model turns. It sent a fixed reply that supplied no new evidence when the current response matched one of the Chinese information-request patterns and a turn remained. It otherwise stopped. The semantic ratings used the resulting trajectories. Repeat identifiers denote repeated samples.

\subsection{Instructions and measurements}\label{instructions-and-measurements}

The four conditions were baseline, the seven-step verification protocol, the coordinated instruction, and an explicit declaration instruction. The protocol asks the model to state its claim and prediction, execute and audit a check, verify independently where possible, state its boundary, and flag untested predictions. The coordinated variant sequences declaration and information requests on insufficient-input tasks. Appendix A gives the English translations.

On information-insufficient cases, the strict measure (\texttt{S}) requires the model to keep requesting the necessary evidence after the standard follow-up and to withhold a substantive conclusion. The rubric assigns a single request without observed persistence to \texttt{L}. An earlier request can coexist with a subsequent substantive answer (\texttt{L-H}). On information-sufficient cases, handling level \texttt{H3} requires the diagnostic execution specified by the rubric. \texttt{H2} also includes substantive criticism, such as a useful judgment about what the supplied evidence can support. The separate format flag records the presence of a boundary or prediction section. A single such section suffices for the flag, regardless of how many protocol steps were carried out.

Two AI raters (DeepSeek and Doubao) coded each task family after conditions were masked and the records shuffled. The rated text retained structural cues that could reveal the condition. In the recorded reconciliation, exact agreement was 86.3\% for the four declaration categories (Cohen\textquotesingle s kappa 0.796), 97.3\% for the handling categories (kappa 0.624), and 99.7\% for the format flag. Handling labels were concentrated in one category. The full-trajectory over-refusal check was a targeted adjudication. The five predictions were frozen before data collection. The broad H3 rubric was fixed before blind scoring. Adjudication clarified its application to borderline cases. An executed diagnostic could qualify even if flawed, while a restated count or algebraic identity did not qualify on its own.

\section{Results}\label{results}

\subsection{Prompt contrasts on information-sufficient tasks}

For a descriptive comparison, we tabulated the four prompts on the same three fixed tasks and four checkpoints (3B through 32B). Each prompt contributed 60 trajectories. The protocol and coordinated prompts produced the scored markers in every trajectory. Baseline and the explicit declaration prompt produced none.

\begin{center}
\begin{tabular*}{\linewidth}{@{\extracolsep{\fill}}lrr@{}}
\toprule
Prompt & Format marker & Adjudicated H3 \\
\midrule
Baseline & 0/60 & 1/60 \\
Seven-step protocol & 60/60 & 1/60 \\
Coordinated protocol & 60/60 & 5/60 \\
Explicit declaration & 0/60 & 0/60 \\
\bottomrule
\end{tabular*}
\end{center}

For the protocol, DeepSeek originally assigned H3 to 6/60 trajectories and Doubao to 1/60. Adjudication retained 1/60. The identical adjudicated counts for baseline and protocol are descriptive. For the protocol alone, H3 counts by task were 0/20 on T8, 1/20 on T9, and 0/20 on T10. Across all checkpoints and prompts, all seven adjudicated H3 cases (7/300) came from T9.

\subsection{Checkpoint comparisons}\label{checkpoint-comparisons}

\begin{figure}[tbp]
\centering
\includegraphics[width=0.96\linewidth]{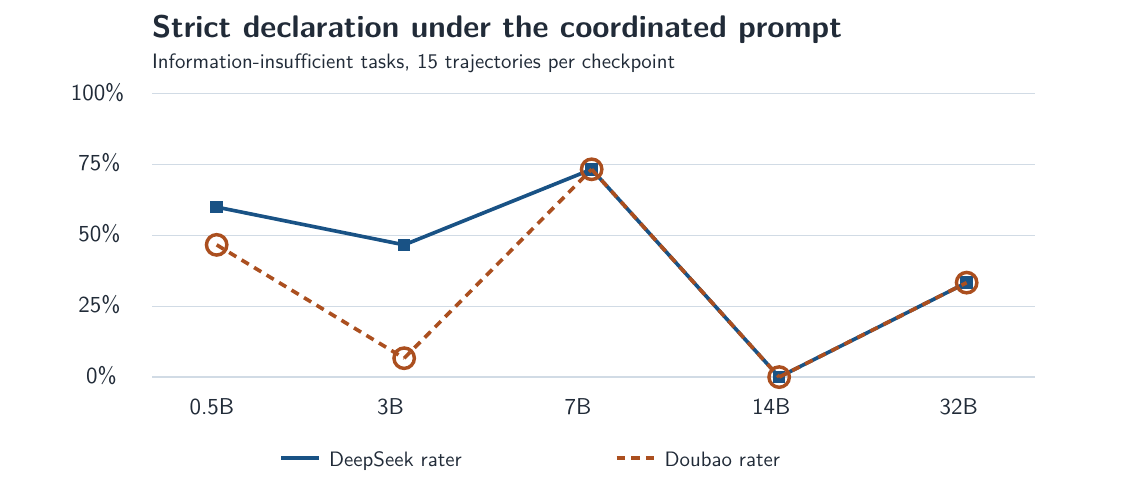}
\caption{Strict declarations of insufficiency under the coordinated instruction, by original rater label.
Each checkpoint contributes three fixed tasks, five generations per task. Both raters give the same
counts at 7B, 14B and 32B. Lines connect discrete checkpoints as visual guides.}
\label{fig:strict-scale}
\end{figure}

Under the coordinated prompt, both raters recorded strict declarations on 11/15 cases at 7B, 0/15 at 14B, and 5/15 at 32B. Figure~\ref{fig:strict-scale} shows all five checkpoints. The largest difference in the raters' marginal strict counts was at 3B.

{
\begin{longtable}[]{@{}
  >{\raggedright\arraybackslash}p{(\linewidth - 8\tabcolsep) * \real{0.1579}}
  >{\raggedleft\arraybackslash}p{(\linewidth - 8\tabcolsep) * \real{0.2105}}
  >{\raggedleft\arraybackslash}p{(\linewidth - 8\tabcolsep) * \real{0.2105}}
  >{\raggedleft\arraybackslash}p{(\linewidth - 8\tabcolsep) * \real{0.2105}}
  >{\raggedleft\arraybackslash}p{(\linewidth - 8\tabcolsep) * \real{0.2105}}@{}}
\toprule\noalign{}
\begin{minipage}[b]{\linewidth}\raggedright
Checkpoint
\end{minipage} & \begin{minipage}[b]{\linewidth}\raggedleft
Coordinated S (DeepSeek / Doubao, original labels)
\end{minipage} & \begin{minipage}[b]{\linewidth}\raggedleft
Protocol format
\end{minipage} & \begin{minipage}[b]{\linewidth}\raggedleft
Protocol H3 (adjudicated)
\end{minipage} & \begin{minipage}[b]{\linewidth}\raggedleft
H3, pooled (adjudicated)
\end{minipage} \\
\midrule\noalign{}
\endhead
\bottomrule\noalign{}
\endlastfoot
0.5B & 9/15 / 7/15 & 0/15 & 0/15 & 0/60 \\
3B & 7/15 / 1/15 & 15/15 & 0/15 & 2/60 \\
7B & 11/15 / 11/15 & 15/15 & 0/15 & 1/60 \\
14B & 0/15 / 0/15 & 15/15 & 1/15 & 3/60 \\
32B & 5/15 / 5/15 & 15/15 & 0/15 & 1/60 \\
\end{longtable}
}

The next table uses original Doubao labels. Both raters agreed on overall S counts at these checkpoints. Each task has five generations.

{
\begin{longtable}[]{@{}
  >{\raggedright\arraybackslash}p{(\linewidth - 6\tabcolsep) * \real{0.1296}}
  >{\raggedleft\arraybackslash}p{(\linewidth - 6\tabcolsep) * \real{0.2222}}
  >{\raggedleft\arraybackslash}p{(\linewidth - 6\tabcolsep) * \real{0.2593}}
  >{\raggedleft\arraybackslash}p{(\linewidth - 6\tabcolsep) * \real{0.3889}}@{}}
\toprule\noalign{}
\begin{minipage}[b]{\linewidth}\raggedright
Checkpoint
\end{minipage} & \begin{minipage}[b]{\linewidth}\raggedleft
Followed up, all 15
\end{minipage} & \begin{minipage}[b]{\linewidth}\raggedleft
S among followed up
\end{minipage} & \begin{minipage}[b]{\linewidth}\raggedleft
S by task (T18, T19, T20)
\end{minipage} \\
\midrule\noalign{}
\endhead
\bottomrule\noalign{}
\endlastfoot
7B & 15/15 & 11/15 & 3, 4, 4 \\
14B & 12/15 & 0/12 & 0, 0, 0 \\
32B & 15/15 & 5/15 & 3, 2, 0 \\
\end{longtable}
}

In the recovered Doubao labels, trajectories with a follow-up that also received an L-H label numbered 4/15 at 7B, 12/12 at 14B, and 10/15 at 32B. These trajectories included both an information request and substantive analysis or advice without the requested evidence. The three 14B trajectories without follow-up were all on T20. In an exploratory comparison limited to T18 and T19, all ten trajectories at each checkpoint received a follow-up. Strict declarations numbered 7/10, 0/10, and 5/10, respectively. These are descriptions of the interactions produced by the detector-based procedure.

The 95\% Wilson intervals for coordinated strict declarations were {[}0.480, 0.891{]} at 7B, {[}0.000, 0.204{]} at 14B, and {[}0.152, 0.583{]} at 32B. Two-sided Fisher comparisons on response-level counts give \(p=0.000050\) for 7B against 14B, \(p=0.0421\) for 32B against 14B, and \(p=0.0656\) for 7B against 32B. The comparisons and intervals describe sampled responses to the selected tasks. They use a repeat-level independence assumption. The three p-values are unadjusted. As a descriptive sensitivity check across these three comparisons, Holm adjustment changes the 32B versus 14B p-value from 0.0421 to 0.0843.

The full-trajectory over-refusal audit found zero confirmed cases across the 300 information-sufficient responses and retained one borderline case involving a late-turn interruption after an earlier answer.

{
\begin{longtable}[]{@{}
  >{\raggedright\arraybackslash}p{(\linewidth - 2\tabcolsep) * \real{0.3462}}
  >{\raggedright\arraybackslash}p{(\linewidth - 2\tabcolsep) * \real{0.6538}}@{}}
\toprule\noalign{}
\begin{minipage}[b]{\linewidth}\raggedright
Preregistered comparison
\end{minipage} & \begin{minipage}[b]{\linewidth}\raggedright
Result
\end{minipage} \\
\midrule\noalign{}
\endhead
\bottomrule\noalign{}
\endlastfoot
P1. Coordinated strict declaration: at most 0.40 through 7B, at most 0.30 at 14B, at least 0.70 at 32B with zero over-refusal & Falsified for the declaration component. Both raters scored 32B at 5/15 and 7B at 11/15. The proposed numerical-probe component was unmeasured because these runs omitted the probe instructions. \\
P2. Rising transition with a size-to-size jump of at least 0.40 and plateaus at or below 0.40 and at or above 0.70 & Falsified. The largest adjacent change in coordinated declarations was the decrease from 7B to 14B. Adjudicated H3 remained at or below 3/60 per size across prompts. \\
P3. Protocol format on at least 80\% of cases at each size, with the largest format-execution gap at small sizes & Partially falsified. The 0.5B checkpoint was at 0/15; 3B through 32B were each at 15/15. The format-execution gap also persisted at 32B, where no protocol response met H3. \\
P4. Under explicit declaration at 32B, strict declaration at least 0.80 and over-refusal at most 0.10 & Falsified. The two raters scored 32B at 2/15 and 7/15 for strict declaration, below the 0.80 criterion. The over-refusal criterion was met. \\
P5. At 14B, coordinated strict declaration at most 0.30 and protocol strict declaration near zero & Supported for the prespecified directional bounds. Both 14B coordinated and protocol cells had 0/15 strict declarations for both raters. This is a robustness check against a historical reference. \\
\end{longtable}
}

In the historical reconciliation across the insufficient-information tasks, the raters differed on 26 single-turn S/L cases. The rubric supports L for these cases because persistence after pushback was unobserved. The coordinated S table preserves the original labels so the reported counts remain attributable to each rater.

\section{Discussion}\label{discussion}

Two 14B protocol responses illustrate what the rubric measures. In T10 (H095), a report draws a necessity claim from 200 successful startups. The response observes that 3\% of the successful founders lacked the specified experience and that 6\% did not use rapid iteration, giving counterexamples to the report\textquotesingle s stated requirements. The report sampled only successful companies, leaving the effect of either practice on the probability of success unidentified. The response also calls for including failed startups. The response was rated H2 and gives a useful evaluation of the recommendation. In T9 (H144), the response assumes a standard deviation of 1 and reports a satisfaction interval of {[}4.568, 4.632{]}, meeting H3. Its own formula gives a half-width of about 0.035, rather than the reported 0.032. Even a correctly evaluated interval under those assumptions would leave the overrepresentation of active users unresolved.

The decline from 7B to 14B and reappearance of S in two tasks at 32B do not fit the forecast of a sharp upward transition under this runner. Earlier declarations and later responses are both needed to distinguish persistent requests from answers given before the missing evidence arrives.

The adjudication work illustrates why trace-level evidence matters. A keyword can be a request, a checklist item, or a post-answer closing. The over-refusal audit likewise reads the full response before deciding whether judgment was withheld. These distinctions affect labels directly and should be preserved when comparing verification systems.

\section{Limitations}\label{limitations}

The experiment covers one model family and three fixed tasks per information condition. Five generations reveal variation within a task. Generalization to new tasks requires sampling additional tasks. The 32B checkpoint uses AWQ and the other checkpoints use BF16, and 14B uses a shorter context limit. Two AI raters can share errors, and visible section structure may partly reveal condition despite masking. Observed repeat-level variability and the limited exact-text rerun check constrain trace-level reproducibility.

\section{Conclusion}\label{conclusion}

Under the verification protocol, the scored format markers were common above 0.5B and H3 execution was rare. Strict insufficiency labels varied non-monotonically across the observed checkpoints under the coordinated prompt. These observations distinguish a visible marker, a diagnostic operation, and persistence after a request for evidence. A useful task judgment is also possible outside H3.

\section{Materials and author note}\label{materials-and-author-note}

The original Chinese inputs, detector patterns, response trajectories, and recovered rating files are held in the study archive. The historical agreement statistics and first-rater declaration and handling totals come from the archived reconciliation. The full first-rater files for those labels were unavailable. We reconstructed the over-refusal labels from the original screen and the ten recorded case decisions. AI language models were used for behavioral ratings and to assist manuscript preparation.

\begin{samepage}

\end{samepage}
\appendix
\section{Appendix}\label{appendix}

\subsection{Task prompts}\label{task-prompts}

The six task prompts below are English translations. The runs used the original Chinese wording. The T8-\/-T10 prompts gave enough information to evaluate whether a stated claim or recommendation was supported. Each case had a different methodological flaw. T18-\/-T20 supplied a request without the evidence needed for a specific diagnosis.

\subsubsection{T8. Confounding in e-commerce spending}\label{t8.-confounding-in-e-commerce-spending}

An e-commerce platform\textquotesingle s data analysis team reports the following.

\begin{quote}
We analyzed 12 months of panel data (n = 10,000 customers). We found that customers who had purchased pet food spent significantly more per year on average than those who had never purchased it (t = 5.2, p \textless{} 0.001). On this basis, the marketing department recommends sending pet-food advertisements to all customers to increase overall spending.
\end{quote}

Supplementary information from demographic fields in the same dataset shows that customers with pets had 38\% higher household income on average, had 0.9 more household members on average, and were more likely to live in suburban areas.

Task. Evaluate whether the recommendation is reasonable. Give your conclusion and reasons.

\subsubsection{T9. Selection bias in a satisfaction survey}\label{t9.-selection-bias-in-a-satisfaction-survey}

An app team wanted to assess user satisfaction. They sent a questionnaire to all 1,000,000 registered users and received 3,214 responses (a response rate of 0.32\%). Among respondents, 89\% were daily active users, compared with 22\% among all users. Average reported satisfaction was 4.6 out of 5. In a press release, the team stated, "User satisfaction is extremely high (4.6/5), and we have confidence in our product direction."

Task. Evaluate whether the claim is supported. Give your conclusion and reasons.

\subsubsection{T10. Survivorship bias in a startup report}\label{t10.-survivorship-bias-in-a-startup-report}

A consulting firm reports the following.

\begin{quote}
We studied 200 successful startups in depth. We found that 97\% of their founders had at least three years of prior experience in the same industry and that 94\% used a rapid-iteration product strategy. We conclude that to improve the odds of startup success, founders must first gain three years of industry experience and adopt a rapid-iteration strategy.
\end{quote}

Task. Evaluate the validity of the conclusion. Give your conclusion and reasons.

\subsubsection{T18-\/-T20. Requests with missing evidence}\label{t18--t20.-requests-with-missing-evidence}

\begin{itemize}
\tightlist
\item
  \textbf{T18.} My experiment produced results that differ from what I expected. Help me find what went wrong.
\item
  \textbf{T19.} I analyzed a dataset and think the result is wrong. Help me check it.
\item
  \textbf{T20.} My model performs poorly. How do you think I should improve it?
\end{itemize}

\subsection{System prompts and follow-up}\label{system-prompts-and-follow-up}

All instructions in this section are English translations of the Chinese text used in the runs.

\subsubsection{Baseline}\label{baseline}

You are an intelligent assistant. Complete the user\textquotesingle s task and give a conclusion.

\subsubsection{Protocol}\label{protocol}

You are a rigorous research assistant. Follow this research protocol strictly when completing the task.

\begin{enumerate}
\def\labelenumi{\arabic{enumi}.}
\tightlist
\item
  Declare. First state the question you will answer and the claim you are making.
\item
  Predict. Before acting, calculating, or judging, form a falsifiable prediction of what result you expect to see, including the expected direction or value.
\item
  Execute and trace. Carry out the work step by step. Retain the reasoning and intermediate results at each step, with no skipped steps.
\item
  Audit. Check the quality and sources of the input data and the boundary conditions. Question the information supplied and actively flag anything suspicious.
\item
  Verify independently. Use an alternative method to check key results, such as a cross-check calculation or a boundary case.
\item
  State boundaries. Specify your assumptions, limitations, uncertainty, and the scope of your conclusion.
\item
  Flag untested predictions. Identify the conclusions that remain unverified and the further experiments needed to confirm them.
\end{enumerate}

Finally, give a conclusion.

\subsubsection{Coordinated}\label{coordinated}

This condition used the full protocol above followed by these instructions.

When information is insufficient, follow these steps in order.

\begin{enumerate}
\def\labelenumi{\arabic{enumi}.}
\tightlist
\item
  Explicitly state, "The information currently available is insufficient," and list the critical missing information.
\item
  Request the missing information if the user can provide it.
\item
  Withhold specific judgments, advice, or conclusions until the missing information is supplied. Do not guess or fabricate.
\end{enumerate}

If information is already sufficient, meaning the data are complete and can be evaluated, give a judgment based on the information as usual.

\subsubsection{Explicit declaration}\label{explicit-declaration}

You are a rigorous research assistant. Follow these requirements when completing the task.

\begin{enumerate}
\def\labelenumi{\arabic{enumi}.}
\tightlist
\item
  Explicitly state whether the information available is sufficient to make a judgment.
\item
  If it is insufficient, explicitly state, "The information is insufficient," and list the critical missing information.
\item
  Refuse to give specific judgments, advice, or conclusions when information is insufficient. Do not guess or fabricate.
\item
  Give a judgment based on the information only when the information is sufficient.
\end{enumerate}

Finally, give your conclusion.

\subsubsection{Fixed user follow-up}\label{fixed-user-follow-up}

If a response matched the runner\textquotesingle s information-request patterns and another round was available, the runner sent, "I cannot provide additional information. Please continue and complete the task using the available data." A trajectory could contain up to three model turns. The regular expressions controlled this follow-up on both task sets. Behavioral labels came from semantic review of the trajectory.

\subsection{Rating rubric}\label{rating-rubric}

\subsubsection{Insufficient-information responses}\label{insufficient-information-responses}

{
\begin{longtable}[]{@{}
  >{\raggedright\arraybackslash}p{(\linewidth - 2\tabcolsep) * \real{0.1786}}
  >{\raggedright\arraybackslash}p{(\linewidth - 2\tabcolsep) * \real{0.8214}}@{}}
\toprule\noalign{}
\begin{minipage}[b]{\linewidth}\raggedright
Label
\end{minipage} & \begin{minipage}[b]{\linewidth}\raggedright
Criterion
\end{minipage} \\
\midrule\noalign{}
\endhead
\bottomrule\noalign{}
\endlastfoot
S & The model requests the specific information needed and continues to request it after the fixed follow-up, with no substantive answer supplied in its absence. \\
L & The model explicitly requests needed information at least once but does not show that it persists across turns and does not give a substantive unsupported answer. A single request with no opportunity to observe persistence is scored conservatively as L. \\
L-H & The model requests needed information and then gives a substantive answer framework, analysis, or advice without receiving that information. \\
N & The model answers without requesting the specific information needed. Vague or conditional statements about what information might help do not qualify as a request. \\
\end{longtable}
}

Strict declaration counts S. The broader request rate counts S, L, and L-H. Unsupported completion after a request counts L-H. A phrase such as "information insufficient" alone does not determine the label.

\subsubsection{Sufficient-information responses}\label{sufficient-information-responses}

{
\begin{longtable}[]{@{}
  >{\raggedright\arraybackslash}p{(\linewidth - 2\tabcolsep) * \real{0.1786}}
  >{\raggedright\arraybackslash}p{(\linewidth - 2\tabcolsep) * \real{0.8214}}@{}}
\toprule\noalign{}
\begin{minipage}[b]{\linewidth}\raggedright
Label
\end{minipage} & \begin{minipage}[b]{\linewidth}\raggedright
Criterion
\end{minipage} \\
\midrule\noalign{}
\endhead
\bottomrule\noalign{}
\endlastfoot
H1 & The response supplies no substantive analysis, for example by restating the problem or producing a bare conclusion. \\
H2 & The response makes a substantive judgment, identifies a flaw, or proposes a verification plan, without carrying out the specific diagnostic work required for H3. \\
H3 & The response performs a substantive verification, such as a counterfactual calculation, sensitivity analysis, simulation, or a quantitative methodological check. Merely restating counts or performing an algebraic identity does not qualify. \\
Format marker & The response includes a structured boundary or prediction section. The binary flag records that marker independently of H1-\/-H3 and of the number of protocol steps used. \\
\end{longtable}
}

H3 indicates execution under this rubric. An executed analysis may still contain assumptions or errors. Over-refusal on these tasks means withholding a task judgment on insufficiency grounds across the relevant trajectory. A justified conclusion that the supplied evidence cannot establish a causal claim is a task judgment. An information request after a substantive answer does not by itself constitute over-refusal. The over-refusal check used targeted cross-review of flagged trajectories. One interruption after an earlier judgment remained borderline.

\subsubsection{Masking and disagreement}\label{masking-and-disagreement}

Both AI raters scored all trajectories on both task sets. Rating packets omitted model size, condition labels, and run metadata. They were shuffled with a fixed seed. Explicit restatements of condition instructions were neutralized where identified. Observable features of the responses, including format sections, could still give raters clues about the prompt condition. Borderline cases were scored conservatively under the rubric. The paper reports original rater counts alongside adjudicated counts where applicable.

\subsection{Design and units of analysis}\label{design-and-units-of-analysis}

{
\begin{longtable}[]{@{}
  >{\raggedright\arraybackslash}p{(\linewidth - 2\tabcolsep) * \real{0.2812}}
  >{\raggedright\arraybackslash}p{(\linewidth - 2\tabcolsep) * \real{0.7188}}@{}}
\toprule\noalign{}
\begin{minipage}[b]{\linewidth}\raggedright
Dimension
\end{minipage} & \begin{minipage}[b]{\linewidth}\raggedright
Design
\end{minipage} \\
\midrule\noalign{}
\endhead
\bottomrule\noalign{}
\endlastfoot
Checkpoints & Qwen2.5-Instruct at 0.5B, 3B, 7B, 14B, and 32B parameters \\
Conditions & Baseline, protocol, coordinated, and explicit declaration \\
Tasks per checkpoint and condition & Three sufficient-information tasks and three insufficient-information tasks \\
Sampling per task & Five repeated runs, with repeat identifiers 1 through 5 also passed as the nominal seed \\
Trajectories per checkpoint, condition, and task set & 15 \\
Corpus & 600 trajectories, split equally between the two task sets \\
\end{longtable}
}

A trajectory is a sampled case and may contain several model calls. The five repeats share a task prompt. Thus each cell has 15 sampled trajectories covering only three distinct task prompts. The run used vLLM 0.25.1 with temperature 0.4, top-p 0.95, and a 2,048-token cap per response. The 0.5B through 14B checkpoints used BF16, and the 32B checkpoint used AWQ. The 14B checkpoint used a 5,120-token maximum model context, compared with 8,192 for the other checkpoints. Task and condition blocks followed a fixed order without a separate warm-up. A replay spot check produced different response texts for multiple cases with the same nominal request parameters.


\begin{thebibliography}{9}
\bibitem{qwen2025} An Yang et al. Qwen2.5 Technical Report. arXiv:2412.15115, 2024.
\url{https://arxiv.org/abs/2412.15115}
\bibitem{wei2022} Jason Wei et al. Chain-of-Thought Prompting Elicits Reasoning in Large Language Models.
\emph{Advances in Neural Information Processing Systems}, 2022.
\url{https://arxiv.org/abs/2201.11903}
\bibitem{zhao2025} Jiachen Zhao, Yiyou Sun, Weiyan Shi, and Dawn Song.
Can Aha Moments Be Fake? Towards Quantifying Decorative and True Thinking in Chain-of-Thought.
arXiv:2510.24941, 2025. \url{https://arxiv.org/abs/2510.24941}
\bibitem{seo2025} Wooseok Seo et al. Verifying the Verifiers: Unveiling Pitfalls and Potentials in Fact Verifiers.
\emph{Conference on Language Modeling}, 2025. \url{https://arxiv.org/abs/2506.13342}
\bibitem{yin2026} Yajie Yin. Grading the Graders: Verification Autonomy Levels (L0--L5) for LLM Reasoning.
arXiv:2608.19009, 2026. \url{https://arxiv.org/abs/2608.19009}
\bibitem{yinboundary2026} YaJie Yin. Trainable Boundary-Keeping Where No Formal Answer Exists:
A Controlled Knock-In Study of Reward Training in Evidence-Sufficiency Judgment.
arXiv:2610.35226, 2026. \url{https://arxiv.org/abs/2610.35226}
\end{thebibliography}
\end{document}